\documentclass[pdflatex,sn-basic]{sn-jnl}

\usepackage{graphicx}%
\usepackage{multirow}%
\usepackage{amsmath,amssymb,amsfonts}%
\usepackage{amsthm}%
\usepackage{mathrsfs}%
\usepackage[title]{appendix}%
\usepackage{xcolor}%
\usepackage{textcomp}%
\usepackage{manyfoot}%
\usepackage{booktabs}%
\usepackage{algorithm}%
\usepackage{algorithmicx}%
\usepackage{algpseudocode}%
\usepackage{listings}%
\usepackage{adjustbox}
\usepackage{tabularx} 
\usepackage{longtable}
\usepackage{subcaption}
\newcommand{\orcidlink}[1]{\href{https://orcid.org/#1}{\texttt{#1}}}

\theoremstyle{thmstyleone}%
\theoremstyle{thmstyletwo}%

\theoremstyle{thmstylethree}%

\begin{document}

\title[Comparative Study of Pretrained Transformers for Quranic ASR]{A Comparative Study of Pretrained Transformer Models for Quranic ASR: Speech Representations, Label Formats, and Dataset Composition}


\author[1]{\fnm{Nabil Mosharraf} \sur{Hossain}}
\author*[1,2]{\fnm{Riasat} \sur{Islam}\,\orcidlink{0000-0002-1419-8068}}\email{riasat.islam@qmul.ac.uk}
\author*[3]{\fnm{Unaizah} \sur{Obaidellah}\,\orcidlink{0000-0003-4822-2174}}\email{unaizah@um.edu.my}

\affil[1]{\orgname{Greentech Apps Foundation}, \country{United Kingdom}}
\affil[2]{\orgname{Queen Mary University of London}, \country{United Kingdom}}
\affil[3]{\orgname{University of Malaya}, \country{Malaysia}}


\abstract{Quran Automatic Speech Recognition (ASR) aims to convert Quranic recitation into text, enabling applications such as aided memorisation tools and Quranic search engines. However, existing ASR models often exhibit high Word Error Rates (WER) on user-recited verses and lack full coverage of the Quranic corpus. This paper presents a systematic empirical study of domain-specific fine-tuning of pretrained Transformer-based models for Quranic ASR, using advanced speech feature extraction methods: Wav2Vec2.0, HuBERT, and XLS-R. These models apply self-supervised learning by masking portions of input audio and using Transformer architectures to learn context-aware speech features. The pretrained models are fine-tuned on a filtered Quranic dataset exceeding 870 hours of professional and user recitations. Through comprehensive ablation studies across feature extractors, output label formats, training strategies, and clip durations, we identify the key factors that affect transcription accuracy in this domain. Our best-performing configuration achieves a WER of 0.08 on the EveryAyah subset and 0.11 on the combined EveryAyah+Tarteel setting, representing roughly a five-percentage-point gain over the Citrinet baseline (WER = 0.163) while reducing combined-model training time from 140 hours to 40 hours. Arabic text without diacritics yields the best fine-tuning results, and Wav2Vec2-XLSR-53 provides the strongest overall representation. Future work includes improving dataset quality and developing phoneme-aware models to extract deeper speech feature representations for Tajweed-sensitive applications.}




\maketitle

\section{Introduction}\label{sec1}

\subsection{Background and Context}
The Arabic language is one of the oldest and most distinguished languages in the world, noted for its originality and adaptability \cite{al2015arabic}. With approximately 290 million native speakers and 132 million non-native speakers, it is the most widely spoken Semitic language \cite{straub2020languages}. It is also one of the six official languages of the United Nations (UN) \cite{kher_arabic_language}. While Modern Standard Arabic (MSA) is used in contemporary communication, Classical Arabic (CA) remains central due to its use in the Qur'an \cite{vanputten2020classical}. The Qur'an is a foundational text in Islam, forming one of the pillars of the Islamic faith and containing divine parables, commands, and teachings \cite{faris2023exploring}.

Recitation and memorization of the Qur'an occupy a central role in Islamic education \cite{talebe2025effect}. It is vital to preserve the Qur'an in its audio form, recited word-by-word as it was revealed to the Prophet Muhammad (peace be upon him), as Muslims believe it to be divine. The rules of recitation, known as \textit{Tajweed}, are critical to ensure accurate pronunciation and proper delivery \cite{elhadj2012tajweed}. The term \textit{Tajweed} derives from the Arabic root ``Jawwada,'' meaning to improve or enhance speech accuracy \cite{alfaries2015tajweed}. However, many Muslims do not speak Arabic as their first language and often struggle to correctly recite verses due to a lack of regular practice and understanding of linguistic nuances \cite{dajani2014difficulties}.

Traditionally, Qur'anic recitation is learned under the supervision of a teacher who listens and provides feedback. However, this method requires significant time investment and face-to-face interaction. Challenges such as environmental distractions, teacher availability, and high student-to-teacher ratios can hinder effective learning \cite{ismail2019retaining}. Therefore, there is a growing need for automated systems that assist learners in recitation and memorization without constant supervision \cite{aziz2019comparison}.

Automatic Speech Recognition (ASR) offers a solution by converting spoken language into text, enabling machines to interpret human speech. ASR has been successfully applied in fields such as education, healthcare, robotics, telecommunications, and customer service \cite{alharbi2021automatic}. In the context of Qur'anic recitation, ASR can support self-learning by detecting pronunciation errors and identifying missing words, thus reducing reliance on human teachers \cite{ibrahim2015problems}.

Despite these advancements, Arabic ASR systems face challenges due to limited labelled data, diverse dialects, and the lack of diacritics in textual data \cite{alrumiah2023intelligent}. Recent Arabic ASR benchmarks such as MGB-2, MGB-3, and MGB-5 have made progress in this domain \cite{ali2018mgb3, ali2019mgb5}. For instance, the MGB-2 dataset contains 1,200 hours of broadcast television content, while MGB-3 and MGB-5 provide extended corpora. These benchmarks have achieved Word Error Rates (WER) of 12.5\%, 27.5\%, and 33.8\% respectively \cite{hussein2022arabic}. WER, defined as the percentage of incorrectly transcribed words relative to the total, is a standard metric used to evaluate ASR performance \cite{ibrahim2015problems}.

However, automatic recognition of Qur'anic recitation presents distinct challenges that remain inadequately addressed in current literature \cite{alrumiah2023intelligent}. Several critical research gaps limit the development of robust solutions in this domain. First, existing approaches demonstrate limited coverage of Tajweed rules and restricted training on narrow subsets of Qur'anic verses \cite{alayyoub2018quranrecitation, ahmad2018tajweed}, failing to capture the comprehensive phonetic variations inherent in Classical Arabic recitation patterns. Second, while foundational datasets such as Tarteel \cite{tarteelml} and fine-tuned architectures including Citrinet models \cite{majumdar2021citrinet} have established baseline performance, the potential of advanced frameworks including Conformer \cite{gulati2020conformer, liu2021improved}, unsupervised speech recognition techniques \cite{baevski2021unsupervised}, and DeepSpeech architectures \cite{al2023building} remains largely unexploited for this specific application domain. Third, current evaluation methodologies inadequately account for the unique acoustic characteristics of Tajweed compliance and the distinction between phonetic accuracy and religious correctness in recitation assessment. Finally, existing research lacks real-time processing systems for error detection, memorization support, and speech-to-search functionalities that serve diverse educational requirements across the global Muslim community \cite{alrumiah2023intelligent}.

\subsection{Proposed Work}

This research addresses the identified gaps through systematic innovation in Qur'anic speech recognition. The study is guided by three primary objectives:

\begin{enumerate}
\item To identify critical acoustic and linguistic parameters affecting accurate Qur'anic recitation transcription
\item To develop an advanced Transformer-based model that significantly reduces WER through optimized deep learning methodologies
\item To evaluate and validate the proposed model's performance against existing baseline approaches
\end{enumerate}

The main contribution of this research is to develop a performant speech recognition model that achieves low WER and low CER while improving transcription accuracy. The key novelties of this work include:

\begin{enumerate}
    \item \textbf{Domain-Specific Adaptation of Transformer-based Models for Qur'anic Speech Recognition:} Systematic fine-tuning and optimization of pretrained end-to-end Transformer architectures for the domain of Qur'anic recitation, constituting a thorough empirical study rather than a new architectural contribution.
    
    \item \textbf{Feature Extraction Comparison:} Systematic evaluation of input features using MFCC, Wav2Vec2, HuBERT, and XLS-R speech representations to identify optimal feature extraction approaches for Qur'anic speech recognition.
    
    \item \textbf{Multi-Format Output Label Analysis:} Comparative analysis of four distinct output label formats (Arabic text, Arabic with Tashkeel, English transliteration, Buckwalter transliteration) to determine the most effective representation for minimizing WER in Qur'anic transcription tasks.
    
    \item \textbf{Training Strategy Evaluation:} Investigation of multiple training approaches including scratch training versus fine-tuning methodologies, dataset composition effects (professional versus layman users), and clip duration impact on model performance, providing systematic insights into optimal training configurations.
    
\end{enumerate}

\begin{figure}[h]
    \centering
    \includegraphics[width=0.8\textwidth]{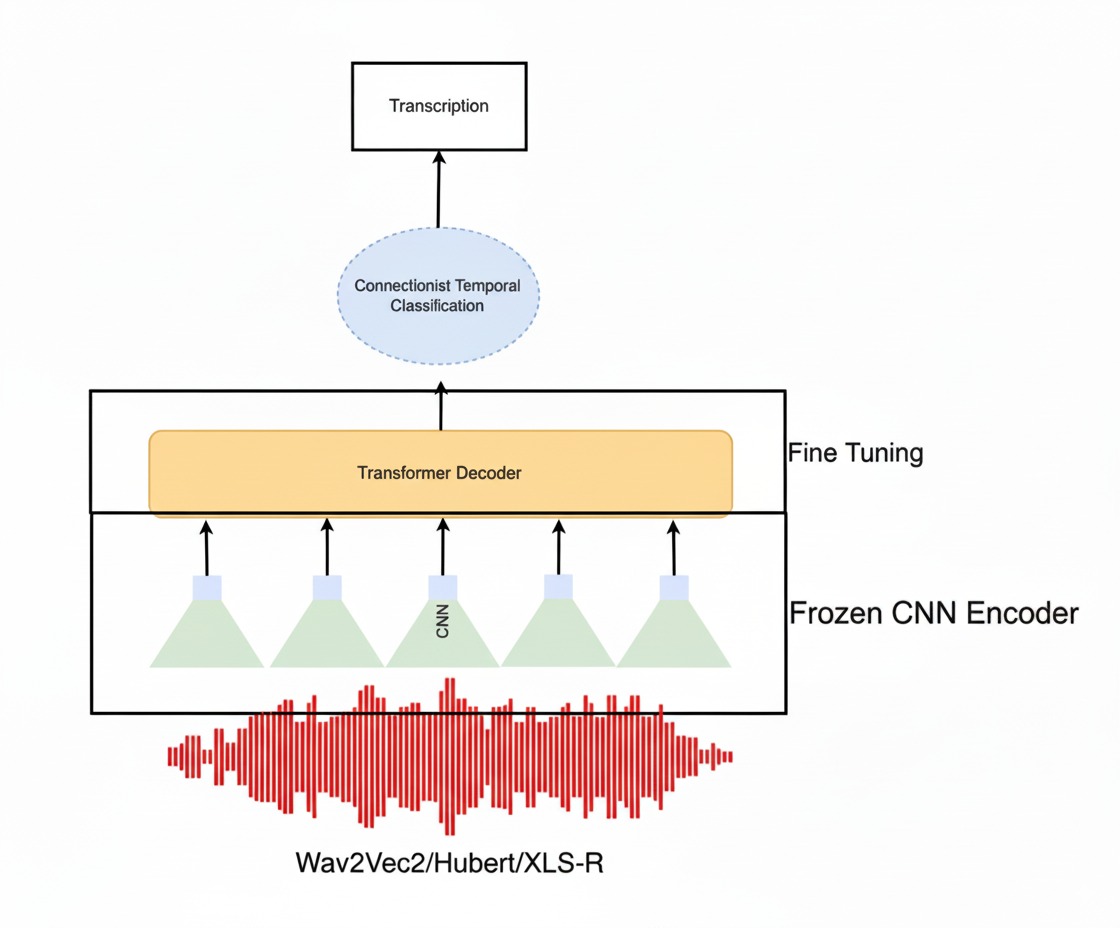}
    \caption{End-to-end Model Architecture: Wav2Vec2/HuBERT/XLS-R with frozen CNN encoder and fine-tuned Transformer decoder using CTC loss for transcription}
    \label{fig:model-architecture}
\end{figure}

Figure~\ref{fig:model-architecture} illustrates the proposed end-to-end model architecture, which utilizes Wav2Vec2/HuBERT/XLS-R with a frozen CNN encoder and a fine-tuned Transformer decoder using CTC loss for transcription.

The research methodology is organized into three phases that address the identified research gaps:

\begin{enumerate}
    \item \textbf{Data Collection and Preprocessing:} This phase involves the collection of audio data, extraction of audio features, and pre-training using Transformer-based architectures.
    
    \item \textbf{Model Training and Fine-Tuning:} In this phase, the model is trained and fine-tuned using various hyperparameter configurations and pretrained models to optimize performance.
    
    \item \textbf{Evaluation and Validation:} The final phase includes evaluating and validating the model against benchmark systems. Performance is assessed using standard metrics such as Word Error Rate (WER) and Character Error Rate (CER).
\end{enumerate}

The proposed research has significant potential to benefit both existing students and introduce new readers and memorizers to Qur'anic study. The developed model enables voice search functionality, allowing users to recite verses and locate their positions within the Qur'an. It supports memorization enhancement by enabling device-based recitation testing and error identification. Additionally, the system can generate subtitles for different surahs, providing word-by-word or ayah-by-ayah guidance for improved learning and understanding. This research facilitates the development of advanced user assessment techniques and human learning enhancement capabilities, including the creation of comprehensive error compendia for hafiz students and exploration of age-related recitation error patterns through data analytics, ultimately transforming Qur'anic study and understanding.

\section{Background}

Automatic Speech Recognition (ASR) has seen significant advancements in recent years, particularly through the application of deep learning and end-to-end architectures \cite{o2024trends}. However, Arabic and Quranic ASR remain uniquely challenging due to linguistic complexity, dialectal diversity, and Tajweed rules \cite{ibrahim2015problems}. This section reviews the literature according to the three primary research objectives of this study.

\subsection{Acoustic and Linguistic Parameters in Arabic and Qur'anic Speech Recognition}

Arabic presents several distinct challenges for ASR systems: its consonantal nature, dialectal diversity, complex morphology, and similar phoneme articulation. Classical Arabic used in the Quran intensifies these challenges due to Tajweed rules that must be followed to ensure accurate pronunciation \cite{elhadj2012tajweed, alfaries2015tajweed}. Additionally, the variability in pronunciation among professional and layman reciters introduces noise that affects model accuracy \cite{dajani2014difficulties}.

Several studies have proposed systems for detecting Tajweed errors \cite{Ismail2014MFCCVQ, tabbaa2015computer, maqsood2016mispronunciation, nazir2019mispronunciation, Alagrami2020Smartajweed, Ziafat2021}. Techniques range from MFCC-VQ to DCNN and BLSTM. However, these are often limited to specific rules or small datasets, highlighting the need for comprehensive acoustic feature identification that captures the full spectrum of Tajweed phonetic variations.

\subsection{Advanced Deep Learning Architectures for Speech Recognition}

Initial automatic speech recognition (ASR) systems were based on Gaussian Mixture Model - Hidden Markov Model (GMM-HMM) architectures. The hybrid Hidden Markov Model - Deep Neural Network (HMM-DNN) model, as proposed by Dahl et al. \cite{dahl2012context}, replaced Gaussian Mixture Models (GMMs) with Deep Neural Networks (DNNs), resulting in significant performance improvements. Subsequent advancements introduced Time Delay Neural Networks (TDNN), Bidirectional Long Short-Term Memory (BLSTM) networks, and optimization techniques such as Lattice-Free Maximum Mutual Information (LF-MMI) \cite{peddinti2015tdnn, khurana2017qats, smit2018aalto}. However, these modular systems tend to be complex, computationally intensive, and less suitable for deployment on mobile and resource-constrained devices.

End-to-end (E2E) models simplify the overall architecture by directly mapping raw audio input to corresponding textual output, thereby improving both recognition performance and training efficiency. The main types of end-to-end models include: (1) Connectionist Temporal Classification (CTC) \cite{graves2014endtoend}, which aligns input and output sequences without requiring pre-segmented data; (2) attention-based Sequence-to-Sequence (Seq2Seq) models \cite{xiong2017humanparity}, which utilize attention mechanisms to dynamically focus on relevant parts of the input sequence during decoding; and (3) Recurrent Neural Network Transducer (RNN-T) models \cite{graves2012sequence}, which combine acoustic modeling and language modeling into a single unified framework. Additionally, multitask learning architectures that integrate both CTC and attention mechanisms \cite{watanabe2017hybrid} have been shown to further enhance model accuracy by leveraging complementary learning signals from both approaches.

Recent innovations in self-supervised speech representation learning include Wav2Vec (Waveform to Vector) \cite{schneider2019wav2vec}, Wav2Vec 2.0 \cite{baevski2020wav2vec}, Cross-Lingual Speech Representations (XLS-R) \cite{babu2021xlsr}, and Hidden-Unit BERT (HuBERT) \cite{hsu2021hubert}. These methods use large volumes of unlabelled audio data to learn generalizable speech features, thereby reducing WER even when only limited labelled data is available.

\subsection{Evaluation Frameworks and Existing Qur'anic ASR Systems}

Performance is commonly evaluated using WER and CER, derived from Levenshtein distance \cite{levenshtein1966binary}. These metrics quantify substitution, insertion, and deletion errors, helping to compare ASR systems effectively.

Quranic ASR research has progressed from traditional GMM-HMM systems \cite{ibrahim2013automated, ould2014phoneme, Amrani2016} to more recent end-to-end and Transformer-based models \cite{ahmad2018tajweed, hussein2022arabic}. However, many models focus on a limited set of verses or reciters. Tarteel.io employed Citrinet-based models using user-generated data \cite{tarteelml}. Prior studies often lack support for layman reciters and fail to generalize across the full Quran \cite{alrumiah2023intelligent}. A detailed summary of recent Quranic ASR systems, including their methods, data sources, and research gaps, is presented in Table \ref{tab:lit-summary}.

\begingroup
\small
\setlength{\tabcolsep}{3pt}
\begin{longtable}{@{}p{0.14\linewidth}p{0.19\linewidth}p{0.16\linewidth}p{0.25\linewidth}p{0.16\linewidth}@{}}
\caption{Summary of Recent Quran Speech Recognition Research}
\label{tab:lit-summary}\\
\toprule
\textbf{Author} & \textbf{Technique (Extraction + Model)} & \textbf{Data} & \textbf{Key Insights (Tajweed + Highlights)} & \textbf{Research Gap} \\
\midrule
\endfirsthead
\caption[]{Summary of Recent Quran Speech Recognition Research (continued)}\\
\toprule
\textbf{Author} & \textbf{Technique (Extraction + Model)} & \textbf{Data} & \textbf{Key Insights (Tajweed + Highlights)} & \textbf{Research Gap} \\
\midrule
\endhead
\midrule
\multicolumn{5}{r}{Continued on next page}\\
\midrule
\endfoot
\bottomrule
\endlastfoot
Ibrahim et al., 2013 & MFCC, GMM + HMM + CMU Sphinx4 & Surah 1 & None; Accuracy 86.41\% & Limited Data \\
Ismail et al., 2014 & MFCC-VQ, GMM + HMM & Surah 114 & Qalqalah; Faster inference, lower accuracy & Limited Data \\
Ould et al., 2014 & MFCC, GMM + HMM & 8 Hours, Juz Amma & None; Accuracy 92\% & Limited Data \\
Tabbaa \& Soudan, 2015 & MFCC, GMM + HMM + SVM & 7.5 hrs, Pro/Layman speakers & 'R' phoneme; Accuracy 91.2\% & Focused on one rule \\
El Amrani et al., 2016 & MFCC, GMM + HMM & 8 hrs, 20 reciters, Surah 1,112-114 & None; 50\% WER & Missing phonemes \\
Yuwan \& Lestari, 2016 & MFCC, GMM + HMM & 180 phonetically rich verses & None; Qscript, WER 22.42\% & WER improvable \\
Maqsood et al., 2016 & MFCC, GMM + HMM + SVM & Non-native speakers & 5 complex letters; Accuracy 97.5\% & Only 5 phonemes \\
AbdulQader Al-Bakeri, 2017 & MFCC, GMM + HMM & Surah 55,114 & None; WER 89.47\% & Limited Data \\
Belkasmi et al., 2017 & MFCC, GMM + HMM + Phoneme duration & 8 hrs, 10 reciters & Phoneme duration; Inconclusive classification & Rule mismatch \\
Ridwan \& Lestari, 2018 & MFCC, GMM + HMM & Full Quran & None; WER 18.53\% & No Tajweed check \\
Mohammed et al., 2018 & MFCC, GMM + HMM & 10 reciters, 600 words & Long Maad, Ghunnah; Duration-based detection & Rule mismatch \\
Ahmad et al., 2018 & MFCC, ANN & 2 reciters, 300 words & Noon Sakinah, Tanween; 77.7\% Accuracy & Limited rules \\
Al-Marri et al., 2018 & MFCC, DNN + HMM & 83 hrs, 100 speakers & Mispronounced letters; 90\% Accuracy, DNN better than GMM & Letter level only \\
Al-Ayyoub et al., 2018 & MFCC, WPD, HMM-SPL, DNN & 3071 audios, 5M + 5F & 8 Tajweed rules; Accuracy 97.7\% & No recitation feedback \\
Nazir et al., 2019 & MFCC, SVM, KNN, NN & 400 non-native speakers & All letters; NN accuracy 90.1\% & Limited Data \\
Thirafi \& Lestari, 2019 & MFCC, HMM-BLSTM (Kaldi) & 10 reciters, 180 verses & None; Avg WER 4.63\% & No Tajweed check \\
Tarteel.io, 2020 & MFCC, End-to-End Citrinet & 80+ hrs user/pro recitations & None; Mobile optimised, fast & WER improvable \\
Alagrami \& Eljazzar, 2020 & MFCC, DNN + SVM & 657 recordings & Idhgham, Ikhfa, Lam; Accuracy 99\% & Word level only \\
Ziafat et al., 2021 & MFCC, BLSTM, DCNN, AlexNet & Arabic letter recordings & None; Accuracy 95.95\% & Letter level only \\
Al-Issa et al., 2023 & MFCC, DeepSpeech (RNN-ASR) & 257,705 male, 5,744 female audio files & No Tajweed modeled; Best WER 0.406 (male) & Gender imbalance \\
Al Harere \& Al Jallad, 2023 & CNN-BiGRU, CTC & Ar-DAD (37 Surahs, 30 reciters) & None; WER 8.34\%, CER 2.42\% & No full Quran coverage \\
Hadwan et al., 2023 & Mel filterbank, E2E Transformer + RNN/LSTM-LM & 10 hrs, 60 reciters, 16 Surahs & Diacritized Arabic; WER 6.16\%, CER 1.98\% (with LM) & Limited to 16 short Surahs; external LM required \\
\end{longtable}
\endgroup

Despite these advancements, existing literature reveals persistent limitations: inadequate coverage of Tajweed rules \cite{Ismail2014MFCCVQ, tabbaa2015computer}, restricted training on limited verses or specific reciters \cite{ahmad2018tajweed, tarteelml}, and insufficient generalization across diverse recitation styles from professional to layman users \cite{alrumiah2023intelligent}. To address these gaps, this research leverages emerging Transformer-based architectures and self-supervised learning methods (Wav2Vec2, XLS-R, HuBERT) fine-tuned on Quranic data, addressing these limitations and enabling robust, scalable ASR solutions.

\section{Methods}
\begin{figure}[h]
    \centering
    \includegraphics[width=0.8\textwidth]{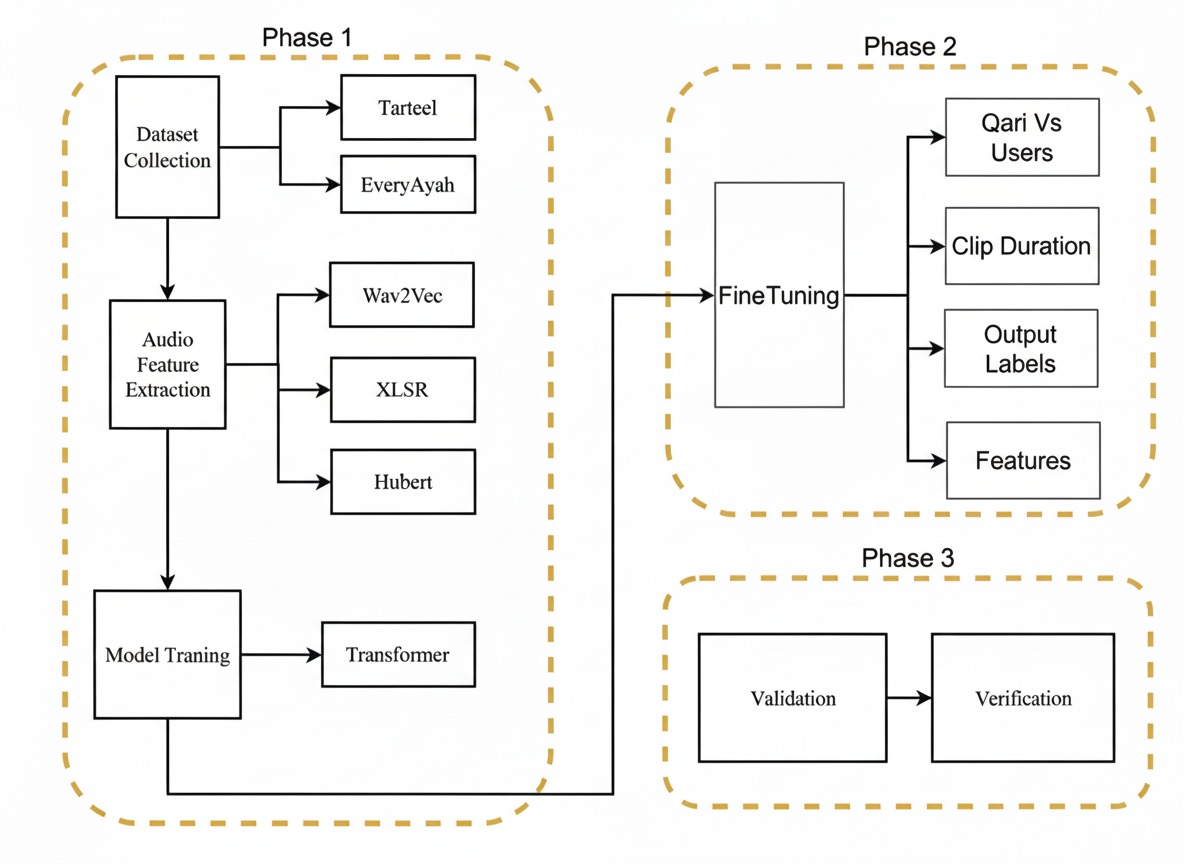}
    \caption{Methodology Flowchart}
    \label{fig:methodology-flowchart}
\end{figure}

Figure~\ref{fig:methodology-flowchart} summarizes the overall workflow proposed in this research. The methodology is structured into three integrated phases addressing data collection and preprocessing, model development and training, and comprehensive evaluation and validation.

\subsection{Data Collection and Preprocessing}

The raw dataset comprises 1310 hours of professional recitations from EveryAyah.com \cite{everyayah} and 62 hours of user-recorded data from Tarteel.io \cite{tarteelml}.

\textbf{EveryAyah Dataset.} This corpus contains recordings from 44 renowned professional male reciters, spanning all 114 Surahs of the Qur'an. Reciters represent internationally recognised recitation schools and were recorded under professional studio conditions with high-quality microphones and minimal background noise. Clip durations range from 1 to 20 seconds with a mean and median of approximately 6 seconds. Transcription annotations are aligned at the verse level using the standard Uthmani Quranic text and were manually verified, providing high annotation quality with full diacritics.

\textbf{Tarteel Dataset.} This corpus consists of user-generated recordings submitted via a mobile application, covering all 114 Surahs. Among the 3,604 recordings with known demographic metadata (out of 20,510 total), approximately 78\% were submitted by male and 22\% by female reciters, spanning a wide range of ages (13 to 56+) and nationalities including Pakistan, Egypt, USA, India, Algeria, Palestine, Bangladesh, and Syria. This geographic and demographic diversity introduces variability in accent, dialect, and recitation proficiency that is representative of real-world usage. Annotation quality is variable: some recordings exhibit background noise, mispronunciation, or minor labelling inconsistencies inherent to crowdsourced data.

These datasets were selected over custom data collection or alternative corpora because they provide comprehensive Qur'an coverage with consistent quality standards, offer authentic user data difficult to replicate in controlled environments, address the research objective of handling both expert and novice reciters, and reduce preprocessing complexity while maintaining semantic coherence.

Audio files were standardized to 16kHz WAV format and filtered to retain clips between 1-30 seconds to manage GPU memory constraints during training. The filtering process reduced the professional dataset from 268,556 clips (1,310 hours) to 230,811 clips (819 hours), and the user dataset from 20,513 clips (62 hours) to 19,771 clips (54 hours). The frequently cited ``over 870 hours'' figure therefore refers to the filtered training corpus rather than the raw recordings. No additional preprocessing such as volume normalization, silence trimming, or noise reduction was applied to preserve natural recitation variations. Silence segments were intentionally retained following recommendations by \cite{baevski2021unsupervised}, as they carry contextual cues important for Qur'anic recitation patterns. The combined dataset was partitioned using an 80:20 train-test split at clip level, stratified to preserve the distribution across chapters and reciters.

\subsection{Feature Extraction}

Pretrained models, namely Wav2Vec2.0 \cite{baevski2020wav2vec}, HuBERT \cite{hsu2021hubert}, and XLS-R \cite{babu2021xlsr}, are used to extract context-aware speech features without silence removal, following \cite{baevski2021unsupervised}. These models are relevant because they provide self-supervised learned representations that capture phonetic variations important for Arabic speech recognition. Wav2Vec2.0 and XLS-R, trained on multilingual datasets, offer cross-lingual capabilities useful for Classical Arabic, and all three models show improved performance in limited-label scenarios typical of Qur'anic datasets. For baseline comparison, MFCC features are also used.

The task involves sequence-to-sequence mapping where extracted audio features are classified to predict character-level transcriptions of Qur'anic recitation. Four distinct output label formats are investigated as part of the research process to identify and compare which output representation yields optimal performance: Arabic without Tashkeel, Arabic with Tashkeel, Buckwalter transliteration, and English transliteration. This comparative analysis enables systematic evaluation of model effectiveness across varying linguistic complexity levels and character set sizes. Output labels are prepared in these four formats as shown in Table~\ref{tab:label-types}. Each output label type was mapped at the character level without additional subword tokenization. SentencePiece tokenization was applied only for the Citrinet baseline model, not the transformer-based models.

\begin{table}[htbp]
\centering
\caption{Different Labeling Formats for Quranic Text, Surah Fatiha, Verse 2 as an Example}
\label{tab:label-types}
\begin{tabular}{lllc}
\toprule
Label Type & Label & Example & Distinct Characters \\
\midrule
Arabic Without Tashkeel & AR & Arabic orthography without diacritics & 39 \\
With Tashkeel & ART & Arabic orthography with full diacritics & 70 \\
Buckwalter Transliteration & BW & AlHmd llh rb AlEAlmyn & 37 \\
Transliteration & TR & Alhamdu lillahi rabbi alAAalameen & 48 \\
\botrule
\end{tabular}
\end{table}

\subsection{Model Architecture and Training}
\begin{figure}[h]
    \centering
    \includegraphics[width=0.8\textwidth]{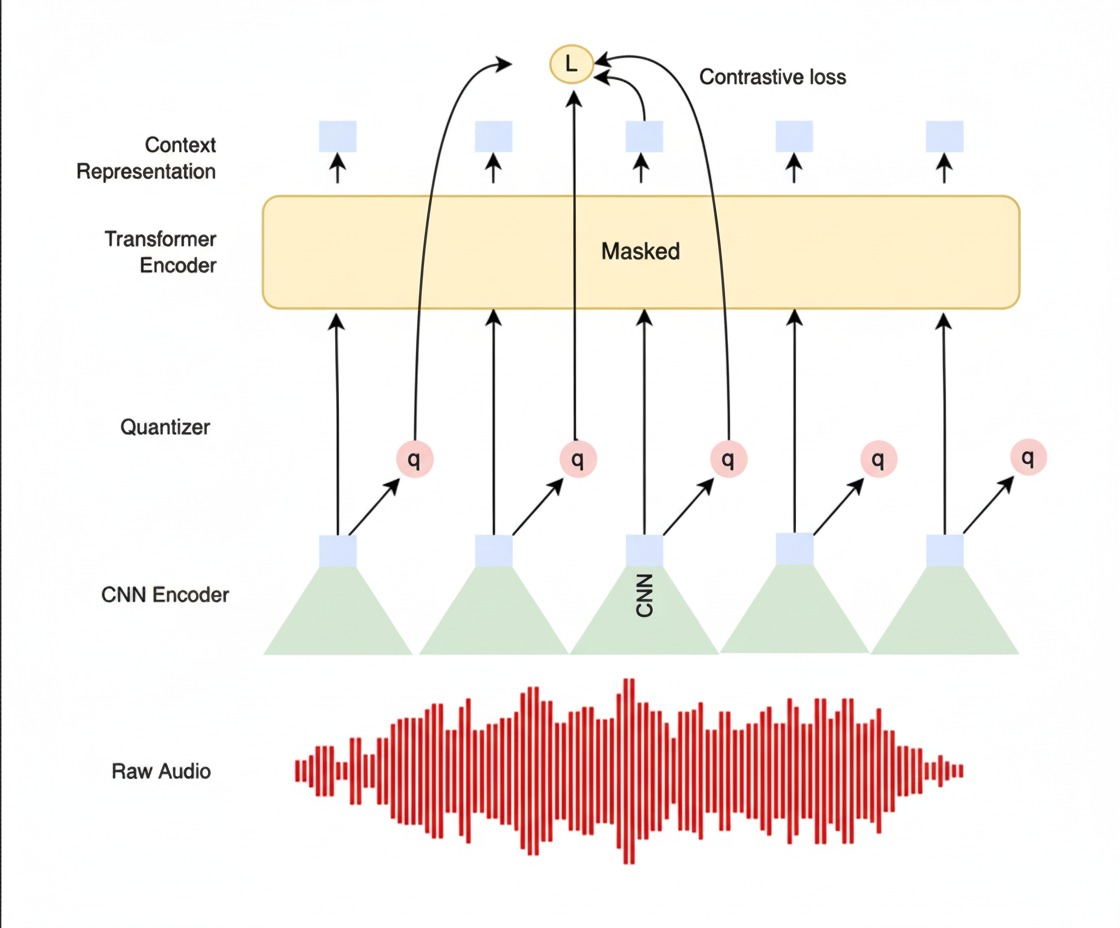}
    \caption{Wav2Vec2.0 ASR Model Architecture}
    \label{fig:wav2vec-architecture}
\end{figure}

Two ASR architectures are employed: a Wav2Vec2.0-based Transformer model (Figure~\ref{fig:wav2vec-architecture}) and a Citrinet-based baseline model \cite{majumdar2021citrinet}. The Transformer model consists of a CNN encoder, quantization module, and a 24-layer Transformer decoder. The actual model used is facebook/wav2vec2-large-xlsr-53.

The architecture illustrated in Figure~\ref{fig:model-architecture} shows the training pipeline with a frozen CNN encoder, Transformer decoder, and Connectionist Temporal Classification (CTC) loss function \cite{graves2014endtoend} for transcription. CTC loss was selected as it enables alignment-free training by allowing the model to learn the mapping between variable-length audio input and output sequences without requiring frame-level alignment. Although greedy decoding was used during training and validation, beam search decoding was also tested during evaluation; however, it did not yield substantial improvements in WER.

The models are fine-tuned using greedy decoding, learning rate of $3 \times 10^{-5}$, dropout of 0.1, and SpecAugment regularization. Batch size is set to 8, with gradient accumulation (steps=3) and mixed-precision training to optimize memory use. The key hyperparameters used during model training are summarized in Table~\ref{tab:model-params}.

\begin{table}[h]
\centering
\caption{Model Parameters}
\label{tab:model-params}
\begin{tabular}{lc}
\toprule
Parameter & Value \\
\midrule
Attention Dropout & 0.1 \\
Hidden Dropout & 0.1 \\
Feature Projection Dropout & 0.0 \\
Mask Time Probability & 0.05 \\
Layer Dropout & 0.1 \\
\botrule
\end{tabular}
\end{table}

\subsection{Evaluation Metrics}

Performance is evaluated using Word Error Rate (WER) and Character Error Rate (CER), which are most suitable for this work due to their complementary strengths in assessing speech recognition accuracy. WER is one of the most common metrics to determine ASR system accuracy \cite{mary2019searching} and is particularly valuable for evaluating speech recognition models because even a single character error in a word substantially increases the error count, providing strict assessment of transcription quality. Character Error Rate (CER) is also evaluated to provide granular assessment of individual character-level accuracy, which is crucial for Qur'anic text where precise character recognition directly impacts Tajweed compliance and religious accuracy.

\textbf{WER:}
\begin{equation}
\text{WER} = \frac{S + D + I}{N}
\end{equation}

\textbf{CER:}
\begin{equation}
\text{CER} = \frac{S + D + I}{C}
\end{equation}
where $S$, $D$, $I$ are substitutions, deletions, insertions, and $N$, $C$ denote total words or characters, respectively.

Training was conducted over approximately 10 to 12 epochs, and model selection was performed based on WER trends on the validation set. Early stopping was considered when performance gains plateaued, to prevent overfitting and reduce training time.

\subsection{Training Environment}

Training was conducted on NVIDIA Tesla P100 (16GB GPU, 25GB RAM) with approximately 300GB of data. Individual Transformer runs typically required 16-17 hours, while the integrated EveryAyah+Tarteel training schedule required about 40 hours. The implementation utilized Google Colaboratory as the primary platform, with Python as the programming language and key libraries including PyTorch for deep learning frameworks, Huggingface for transformer models, SciKit-Learn for machine learning utilities, Numpy for numerical computations, and Librosa for audio processing.

To manage memory constraints, audio was downsampled to 16kHz and training batch size was reduced. Evaluation was performed every 400 steps across 2000 training steps to monitor WER and CER. The dataset was split using an 80:20 ratio for training and testing. The split was performed at clip level while preserving the distribution across chapters and reciters.

\subsection{Baseline Comparison}
The Citrinet baseline, developed by Nvidia, leverages 1D time-channel separable convolutions with SE modules. Tarteel.io fine-tuned Citrinet with MFCC and SentencePiece encoding \cite{kudo2018sentencepiece}. This study compares all models against this baseline to evaluate improvements in performance and scalability for Quranic ASR.

A consolidated overview of the key components used in the methodology is provided in Table~\ref{tab:method-summary}.

\begin{table}[h]
\centering
\caption{Summary of Methodological Components}
\label{tab:method-summary}
\begin{tabular}{ll}
\toprule
Component & Method \\
\midrule
Audio Feature Extractors & Wav2Vec2.0, HuBERT, XLS-R, MFCC \\
Output Labels & Arabic, Tashkeel, Buckwalter, Transliteration \\
Model Architectures & Wav2Vec2 Transformer, Citrinet \\
Loss Function & CTC \\
Decoding & Greedy Search (Beam Search tested) \\
Evaluation Metrics & WER, CER \\
Tools & PyTorch, Huggingface, Google Colab \\
\botrule
\end{tabular}
\end{table}

\section{Results}
\subsection{Training Results}

Following systematic hyperparameter tuning, multiple Automatic Speech Recognition (ASR) models were developed and evaluated. Performance was assessed using two key metrics: Word Error Rate (WER) and Character Error Rate (CER), measured across training steps. To optimize computational efficiency, models were trained separately on professional reciter and user-recited datasets rather than combining both sources, as the distinct characteristics of each dataset type (professional vs. layman recordings) require different optimization strategies for effective convergence.

The following model configurations were investigated:
\begin{itemize}
    \item \textbf{Wav2Vec2:} Trained on professional reciters with Arabic text labels (without diacritics).
    \item \textbf{Tarteel:} Trained on user-recited audio with Arabic text labels.
    \item \textbf{XLS-R:} Trained on professional reciters with Arabic text labels.
    \item \textbf{HuBERT:} Trained on professional reciters with Arabic text labels.
    \item \textbf{Arabic with Tashkeel:} Trained on professional reciters with fully diacritized Arabic text.
    \item \textbf{English Transliteration:} Trained on professional reciters using English transliterated text.
    \item \textbf{Buckwalter Transliteration:} Trained on professional reciters using Buckwalter transliteration.
\end{itemize}

\begin{figure}[htbp]
    \centering
    \begin{subfigure}[b]{0.42\textwidth}
        \includegraphics[width=\textwidth]{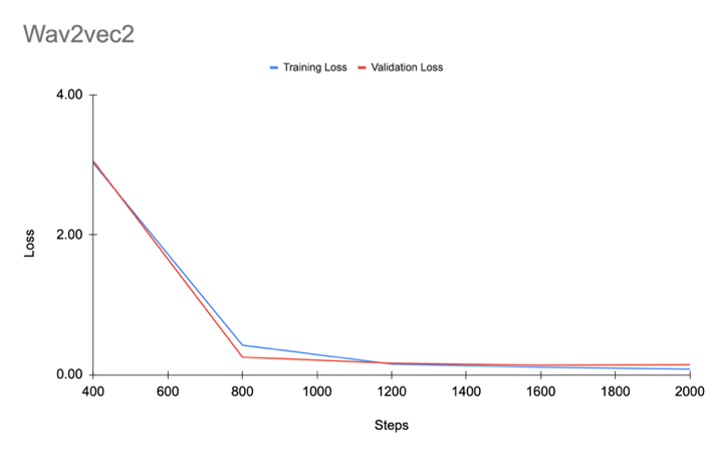}
        \caption{Wav2Vec2}
        \label{fig:loss-a}
    \end{subfigure}
    \hfill
    \begin{subfigure}[b]{0.42\textwidth}
        \includegraphics[width=\textwidth]{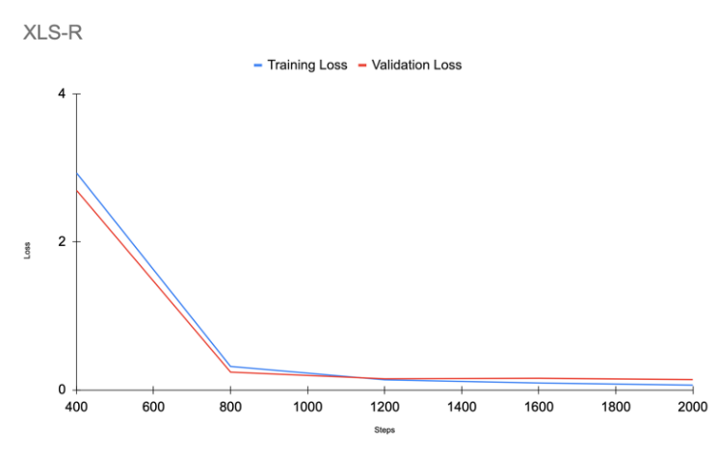}
        \caption{XLS-R}
        \label{fig:loss-b}
    \end{subfigure}

    \vspace{0.5em} 

    \begin{subfigure}[b]{0.42\textwidth}
        \includegraphics[width=\textwidth]{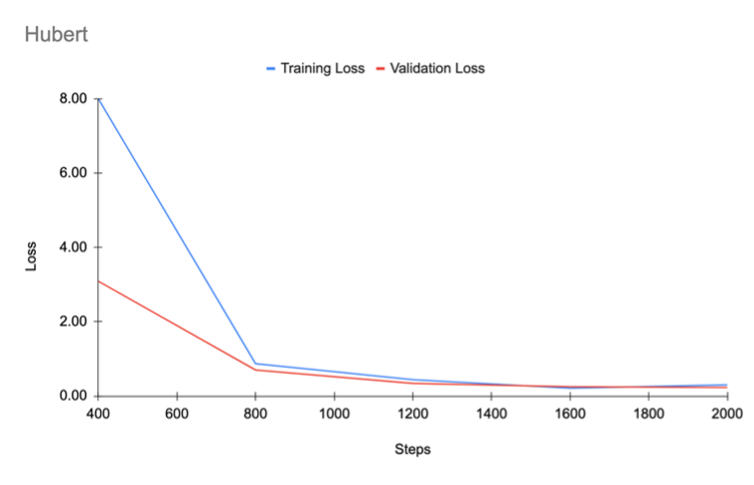}
        \caption{HuBERT}
        \label{fig:loss-c}
    \end{subfigure}
    \hfill
    \begin{subfigure}[b]{0.42\textwidth}
        \includegraphics[width=\textwidth]{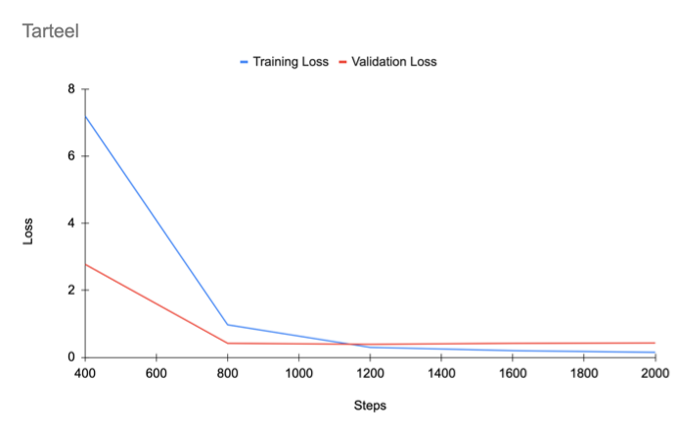}
        \caption{Tarteel}
        \label{fig:loss-d}
    \end{subfigure}

    \vspace{0.5em}

    \begin{subfigure}[b]{0.42\textwidth}
        \includegraphics[width=\textwidth]{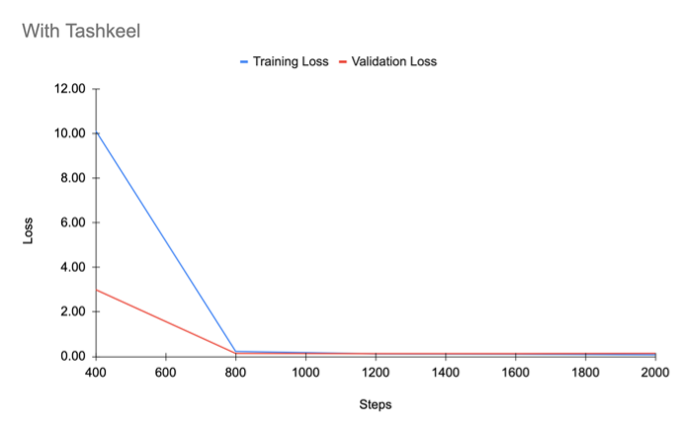}
        \caption{Tashkeel}
        \label{fig:loss-e}
    \end{subfigure}
    \hfill
    \begin{subfigure}[b]{0.42\textwidth}
        \includegraphics[width=\textwidth]{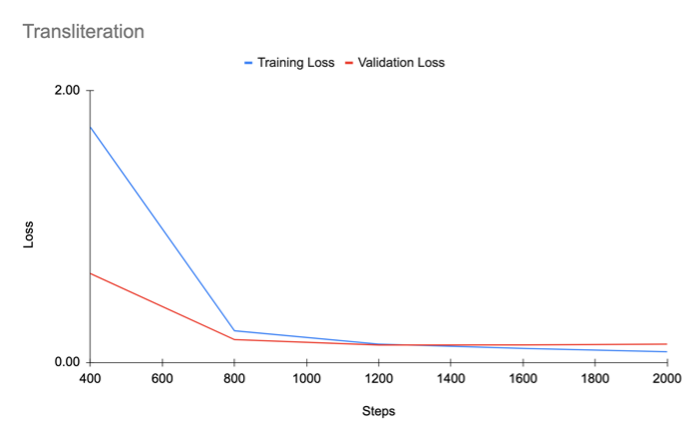}
        \caption{Transliteration}
        \label{fig:loss-f}
    \end{subfigure}

    \vspace{0.5em}

    \begin{subfigure}[b]{0.42\textwidth}
        \centering
        \includegraphics[width=\textwidth]{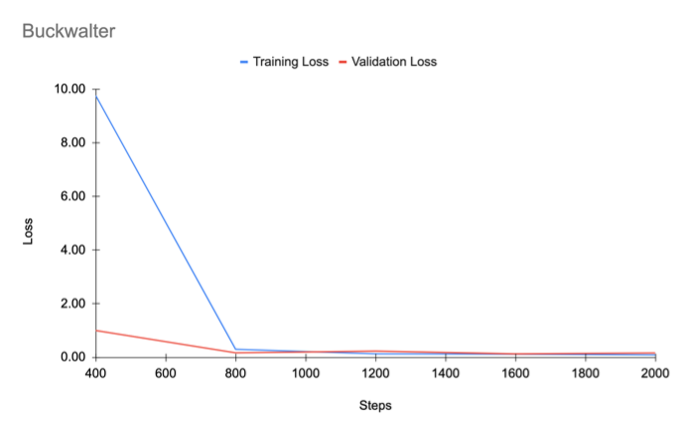}
        \caption{Buckwalter}
        \label{fig:loss-g}
    \end{subfigure}

    \caption{Training vs Validation Loss for Various Models.}
    \label{fig:training-loss}
\end{figure}

\subsection{Training Loss}
Training loss reflects model fitting on training data, while validation loss indicates generalization. Figure~\ref{fig:training-loss} shows loss trends for various models. The Wav2Vec2 model (Figure~\ref{fig:loss-a}) shows rapid convergence with minimal loss gaps, indicating a balanced fit. XLS-R (Figure~\ref{fig:loss-b}) shows slight divergence in the final steps, suggesting mild overfitting. HuBERT (Figure~\ref{fig:loss-c}) exhibits higher training and validation losses, implying underfitting. The Tarteel model (Figure~\ref{fig:loss-d}) and the Tashkeel, Transliteration, and Buckwalter variants (Figures~\ref{fig:loss-e}, \ref{fig:loss-f}, and \ref{fig:loss-g}, respectively) display overfitting behavior, where validation loss diverges from training loss.

\subsection{Character Error Rate (CER)}
\begin{figure}[h]
    \centering
    \includegraphics[width=1\textwidth]{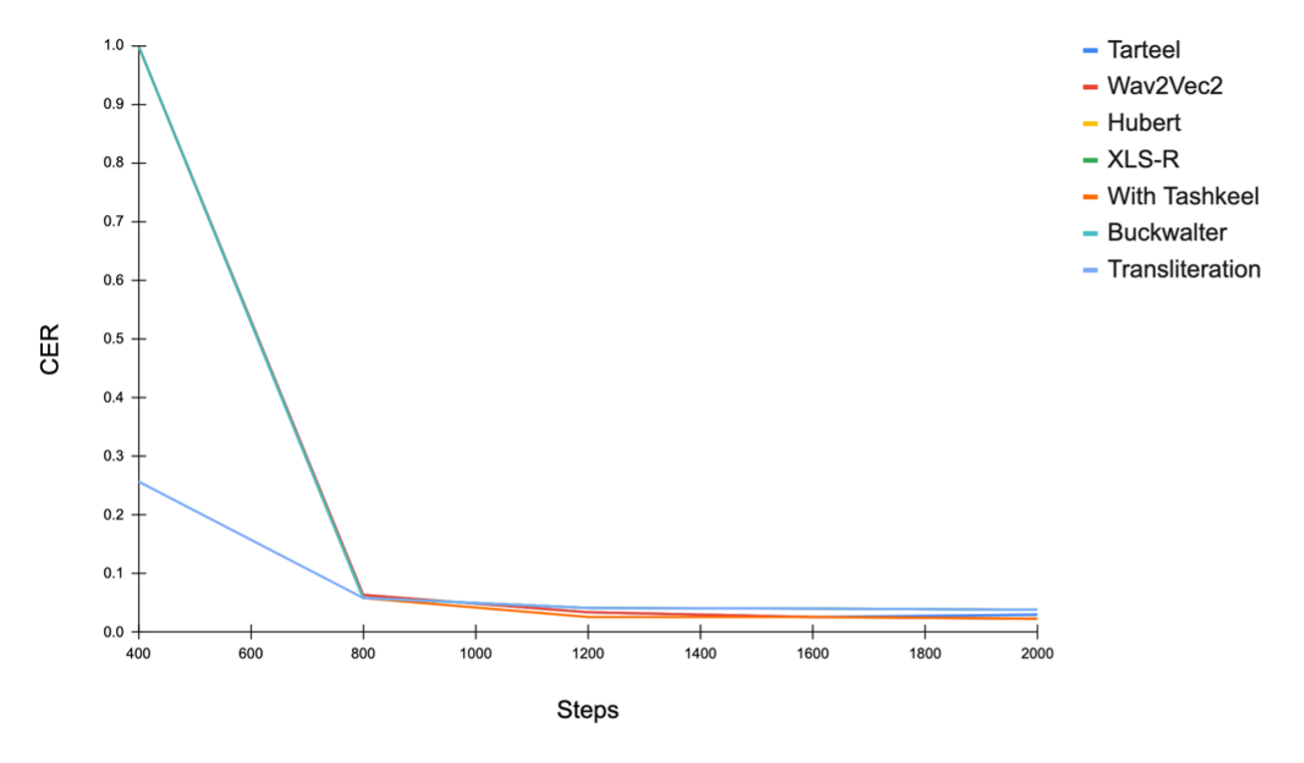}
    \caption{Character Error Rate (CER) comparison for trained models}
    \label{fig:cer-models}
\end{figure}

CER evaluates transcription accuracy at character level. Figure~\ref{fig:cer-models} presents CER trends. All models converged quickly below 0.1. Among the Transformer configurations, Wav2Vec2 achieved the lowest CER, while Tarteel and Tashkeel remained comparatively low. The absolute CER differences are smaller than in the WER comparison, so they should be interpreted as supporting trends rather than the primary discriminator between models.

\subsection{Word Error Rate (WER)}
\begin{figure}[h]
    \centering
    \includegraphics[width=1\textwidth]{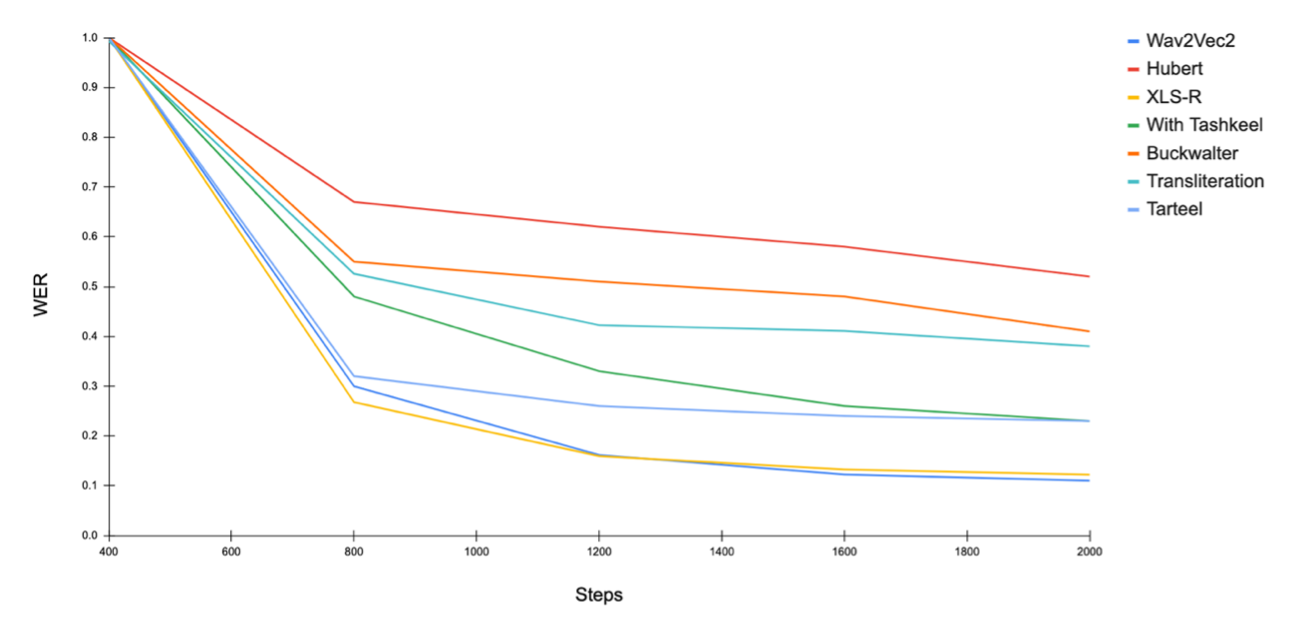}
    \caption{Word Error Rate (WER) comparison for trained models}
    \label{fig:wer-models}
\end{figure}

WER provides stronger differentiation across models. Figure~\ref{fig:wer-models} shows that HuBERT had the highest WER at 0.52. Buckwalter and Transliteration models performed poorly (around 0.40), possibly due to mismatched phonetic representation. The Tarteel-trained model achieved a WER of 0.23, likely impacted by noisy user data, and the Tashkeel model had similar performance. Wav2Vec2 achieved the best result (WER = 0.08), followed closely by XLS-R (WER = 0.09). The Citrinet baseline reported a WER of 0.163 on the combined EveryAyah+Tarteel setting. These findings are further illustrated in Figure~\ref{fig:wer-models} and summarized in Table~\ref{tab:comparison-extended}.

\subsection{Impact on Training Approach}
\begin{figure}[h]
    \centering
    \includegraphics[width=1\textwidth]{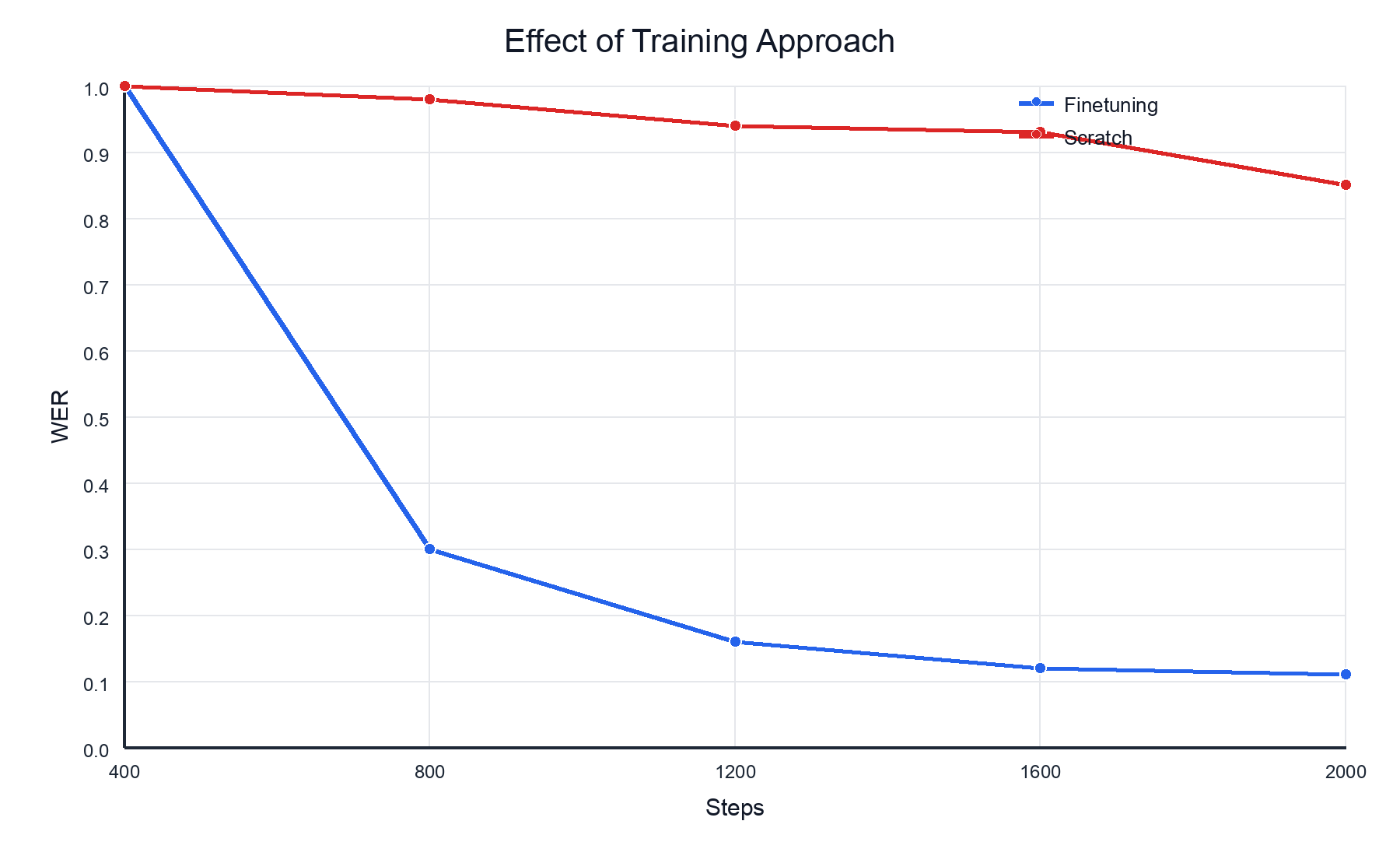}
    \caption{Effect of Training Approach}
    \label{fig:training-approach}
\end{figure}

In this study, the chosen model was Wav2Vec2 pretrained model and the output label was fixed to be Quran Arabic without Tashkeel. Other ablation studies showed that this input feature and output label combination was optimal. Additionally, professional reciter data was chosen. Training from scratch, without any pretrained model, showed that WER slowly decreased to around 0.9 after 2000 steps. However, fine-tuning from pretrained models drastically reduced WER to 0.3 after just 800 steps, and finally to 0.08 after 2000 steps. These results are consistent with prior work \cite{baevski2020wav2vec}. This performance improvement due to different training strategies is depicted in Figure~\ref{fig:training-approach}.

\subsection{Impact of Different Features}
\begin{figure}[h]
    \centering
    \includegraphics[width=1\textwidth]{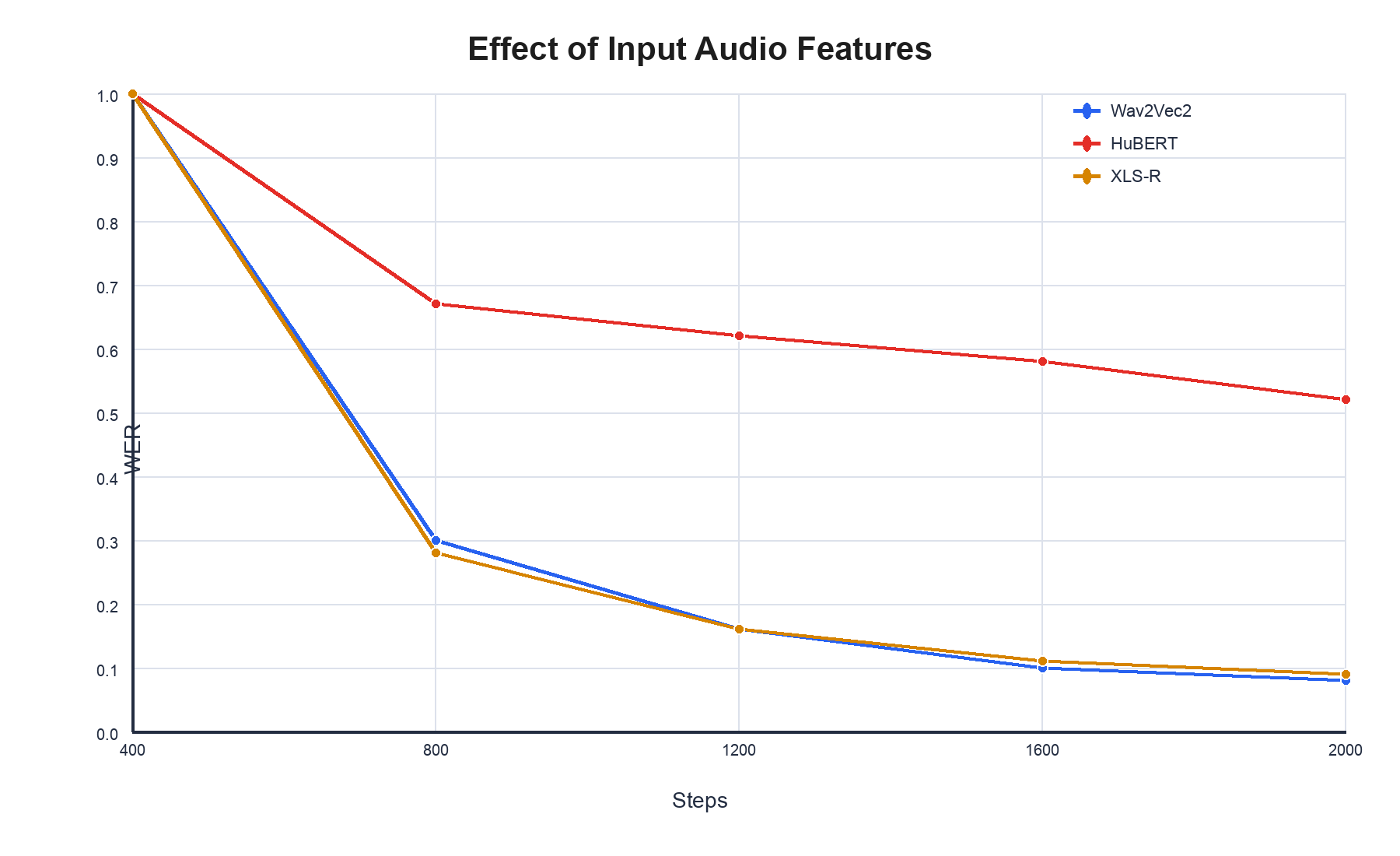}
    \caption{Effect of Input Audio Features}
    \label{fig:feature-extraction}
\end{figure}

Feature extraction is essential in machine learning as it reduces data dimensionality and facilitates more accurate learning. Wav2Vec2-large-XLSR-53, trained on 53 languages, outperformed others, achieving WER = 0.08. HuBERT, trained on English-only data, underperformed (WER = 0.52). XLS-R also performed well (WER = 0.09), indicating the value of multilingual pretraining. The comparative performance of different feature extractors is illustrated in Figure~\ref{fig:feature-extraction}.

\subsection{Impact of Different Output Labels}
\begin{figure}[h]
    \centering
    \includegraphics[width=1\textwidth]{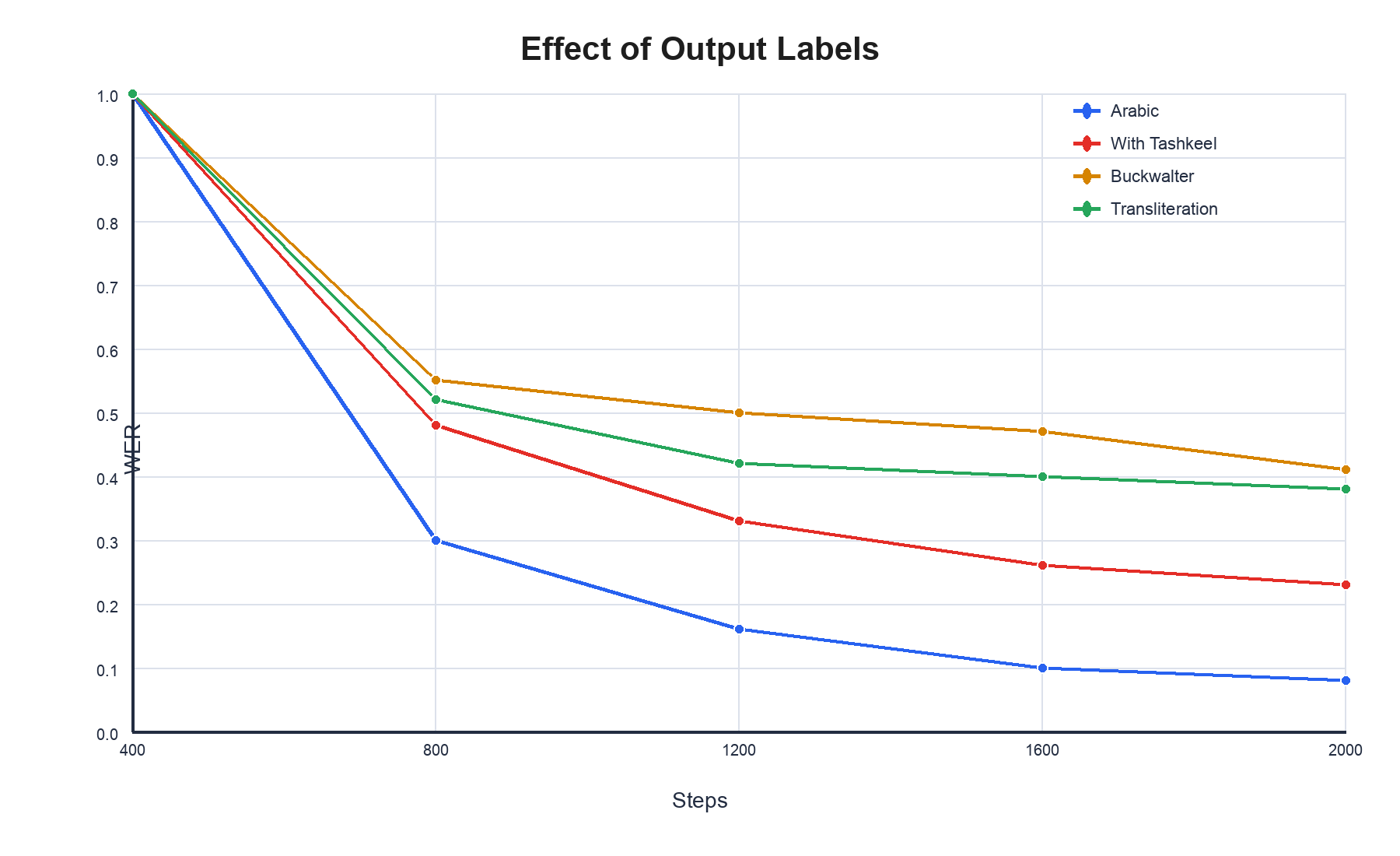}
    \caption{Effect of Output Labels}
    \label{fig:output-labels}
\end{figure}

With Wav2Vec2-large-XLSR-53 as the feature extractor, output labels were compared. Transliteration and Buckwalter remained the weakest settings, ending at WER = 0.38 and 0.41, respectively. Their poor performance may stem from phoneme misrepresentation. The Tashkeel configuration achieved WER = 0.23, while Arabic without Tashkeel remained the best-performing label format with WER = 0.08. These results are visually compared in Figure~\ref{fig:output-labels}.

\subsection{Effect of Dataset Quality and Composition}
Using Wav2Vec2 features and Arabic text without Tashkeel as labels, model training on EveryAyah dataset yielded WER = 0.08. Tarteel dataset, with layman reciters and noisier data, yielded WER = 0.23. Combining datasets (fine-tuning on Tarteel after EveryAyah) resulted in WER = 0.11, indicating generalizability but highlighting the need for improved Tarteel data quality.

It is also important to acknowledge that the Tarteel dataset, while providing diversity in recitations, includes challenges such as inconsistent pronunciation, background noise, and occasional mislabeling. These factors likely contributed to the higher WER observed in models trained solely on Tarteel data and should be addressed in future data preprocessing efforts.

\subsection{Effect of Clip Duration}
Audio clip durations in the dataset varied between 1s to 286s. To evaluate the impact of clip length on model performance and manage GPU memory constraints, experiments were conducted using maximum clip durations of 10s, 20s, 30s, and 40s. Results showed a significant improvement in Word Error Rate (WER) as the clip duration increased. Specifically, the WER was 0.70 for 10-second clips, 0.50 for 20-second clips, and reached its optimal value of 0.075 for 30-second clips. However, using 40-second clips consistently caused out-of-memory errors during training. Therefore, a clip duration of 30 seconds was found to offer the best trade-off between performance and hardware limitations.

\subsection{Common Recognition Errors}

While overall WER and CER metrics indicate high model performance, several recurring transcription errors were observed that reveal specific challenges in Qur'anic speech recognition. The most frequent error category involves confusion between phonetically similar Arabic letters, particularly pairs such as seen and saad, which share similar acoustic properties but differ in emphatic pronunciation crucial for correct Qur'anic recitation. This confusion stems from the subtle articulatory differences that require specialized acoustic modeling to distinguish effectively. Additionally, models demonstrated difficulty recognizing short Ayat (verses) with similar intonation patterns, where brief recitations often lack sufficient contextual information for accurate transcription, leading to misidentification between verses with comparable rhythmic structures. Furthermore, occasional errors occurred in recognizing Ghunnah (nasal sound) and Maad (vowel elongation) pronunciation, which are fundamental Tajweed rules that involve temporal acoustic features extending beyond standard phonemic recognition. These specific error patterns highlight the need for more fine-grained modeling approaches that incorporate Tajweed-aware acoustic features and phoneme-aware decoding mechanisms in future ASR systems designed for Qur'anic recitation.

\subsection{Summary of Model Performance and Baseline Comparison}

The proposed ASR models were evaluated against a baseline Citrinet model, with results summarized in Table~\ref{tab:comparison-extended}. Among all configurations, the Wav2Vec2 model trained on the EveryAyah dataset using Arabic text without Tashkeel achieved the best overall WER, with a value of 0.08 and a corresponding CER of 0.015. Fine-tuning the Tarteel dataset on this pretrained model further improved generalizability, achieving a WER of 0.11, which is roughly a five-percentage-point improvement over the Citrinet baseline (WER = 0.163).

Training time averaged 16 hours for individual models, while the integrated model trained on both datasets required 40 hours. In contrast, the baseline Citrinet model required 140 hours to reach its final performance. These results demonstrate the efficiency and scalability of transfer learning with pretrained Wav2Vec2 models.

In terms of complexity, the proposed Wav2Vec2 model contains approximately 100 million parameters with a model size of 1.2 GB, compared to 30 million parameters and 350 MB for the Citrinet model. Although inference latency is roughly three times slower, the improved transcription accuracy makes the model more suitable for high-accuracy research applications. Future work may explore lightweight variants for mobile deployment.

\begin{table}[h]
\centering
\caption{Comparison of Extended Results with Baseline}
\label{tab:comparison-extended}
\scriptsize
\setlength{\tabcolsep}{2pt}
\renewcommand{\arraystretch}{1.05}
\begin{tabularx}{\textwidth}{@{}>{\raggedright\arraybackslash}p{0.30\textwidth}>{\raggedright\arraybackslash}p{0.12\textwidth}>{\raggedright\arraybackslash}p{0.18\textwidth}>{\raggedright\arraybackslash}p{0.11\textwidth}>{\centering\arraybackslash}p{0.08\textwidth}>{\centering\arraybackslash}p{0.08\textwidth}>{\centering\arraybackslash}p{0.09\textwidth}@{}}
\toprule
Method & Features & Dataset & Labels & WER & CER & Time (hrs) \\
\midrule
Citrinet baseline & MFCC & Combined & Arabic & 0.163 & 0.010 & 140 \\
HuBERT (Arabic) & HuBERT & EveryAyah & Arabic & 0.52 & 0.04 & 16 \\
XLS-R (Arabic) & XLS-R & EveryAyah & Arabic & 0.09 & 0.04 & 16 \\
Wav2Vec2 (Arabic) & Wav2Vec2 & EveryAyah & Arabic & \textbf{0.08} & \textbf{0.015} & 17 \\
Wav2Vec2 (Transliteration) & Wav2Vec2 & EveryAyah & Translit. & 0.38 & 0.023 & 16 \\
Wav2Vec2 (Tashkeel) & Wav2Vec2 & EveryAyah & Tashkeel & 0.23 & 0.021 & 16 \\
Wav2Vec2 (Buckw.) & Wav2Vec2 & EveryAyah & Buckwalter & 0.41 & 0.04 & 16 \\
Wav2Vec2 (Tarteel only) & Wav2Vec2 & Tarteel & Arabic & 0.23 & 0.02 & 16 \\
Wav2Vec2 (Combined) & Wav2Vec2 & Combined & Arabic & 0.11 & 0.017 & 40 \\
\bottomrule
\end{tabularx}
\end{table}

In summary, the results demonstrate that the proposed Wav2Vec2-based ASR model, particularly when trained on professional recitations using Arabic text without diacritics, delivers competitive performance compared to traditional and alternative deep learning approaches in terms of WER and CER. The model also exhibits reduced training time and broader applicability across diverse recitation styles and speaker backgrounds, effectively addressing all three research questions. While some variability in performance was observed across feature extractors, output labels, and dataset sources, the overall findings confirm the effectiveness of self-supervised pretrained models for Quranic speech recognition. These outcomes not only validate the proposed methodology but also provide a strong foundation for future enhancements in model generalization, data quality, and deployment in real-world scenarios. The implications of these findings are further discussed in the next chapter.

\section{Discussion}

This section reflects on the experimental findings, situating them within the broader landscape of Quranic Automatic Speech Recognition (ASR) research. The evaluation of model configurations, dataset compositions, feature representations, and output label formats has yielded valuable insights into the parameters that most significantly impact transcription accuracy.

\subsection{Comparison with Prior Works}

Table~\ref{tab:comparison-otherworks} situates the proposed model among prior Quranic ASR systems. The comparison highlights three dimensions that prior work does not address jointly: (1) \textbf{dataset scale}, as this study uses over 870 filtered hours across professional and user recitations, compared to 8--80 hours in most prior systems; (2) \textbf{robustness to noisy user recitations}, because training and evaluation on Tarteel user data alongside EveryAyah professional recordings allow the model to generalize beyond studio conditions, an aspect absent from systems trained solely on professional reciters; and (3) \textbf{training efficiency}, since self-supervised pretraining reduces required compute from 140 hours (Citrinet baseline) to 16--40 hours, enabling faster iteration. Against MFCC-based GMM-HMM baselines (WER 18--22\%), the performance gain is substantial. The Citrinet model achieves WER 16.3\%, while the proposed Wav2Vec2 model trained on the same combined dataset reaches WER 11.3\%, representing roughly a five-percentage-point gain. These gains are consistent with the broader trend of self-supervised Transformer models outperforming MFCC-driven pipelines \cite{baevski2020wav2vec, hsu2021hubert, babu2021xlsr}. Concurrently, Al Harere and Al Jallad \cite{alharere2023quran} achieved WER 8.34\% using a CNN-BiGRU with CTC on the Ar-DAD dataset (37 Surahs), confirming that end-to-end models deliver strong Quranic ASR performance. Hadwan et al.\ \cite{hadwan2023quran} reported WER 6.16\% and CER 1.98\% using an encoder-decoder Transformer with an external RNN/LSTM language model, trained on 10 hours covering 16 short Surahs. While numerically competitive, this result is evaluated on a narrow, homogeneous subset of the Quran (short Surahs with repetitive phoneme patterns) and relies on an external language model, which makes direct comparison with our full-Quran, LM-free setup impractical. On the EveryAyah-only subset covering the full Quran, our model achieves WER 8.0\%; on the combined filtered corpus exceeding 870 hours, it achieves WER 11.3\%.

A notable finding from Al-Issa et al.~\cite{al2023building} highlights the significant role of transcription quality and domain-specific preprocessing in achieving low WER and CER. Their baseline experiments using MFCC features and Mozilla's DeepSpeech architecture resulted in relatively high error rates, with Word Error Rates (WER) ranging from 46.7\% to 32.1\% in early experiments (Exp-1 to Exp-4). However, after applying progressive cleaning steps such as silence removal, punctuation stripping, fixing corpus alignments, and removing non-Quranic tokens, the performance improved substantially. In Experiment 7, by incorporating a Quran-specific vocabulary set, they achieved a remarkably low WER of 4.6\% and CER of 2.5\%. This demonstrates that ASR performance is not solely dependent on model architecture but also heavily influenced by the quality and consistency of transcription labels. These insights align with our findings, where improved dataset quality and carefully chosen output label formats contributed significantly to accuracy gains. The study underscores the importance of domain-specific preprocessing and offers a compelling direction for future improvements in Quranic ASR systems.

Another relevant benchmark in Quranic ASR research is the DeepSpeech-Quran project by ElDeeb \cite{eldeeb2020deepspeechquran}, which used Mozilla’s DeepSpeech (v0.7.1) RNN-based architecture with MFCC features. The dataset was primarily based on professional recitations from EveryAyah, segmented and labeled at the verse level.

Two configurations were reported:
\begin{itemize}
    \item \textbf{Imam-Only Dataset:} Trained on audio from seven professional reciters. The model achieved WER = 5.6\% and CER = 3.9\%.
    \item \textbf{Imam + Filtered Users Dataset:} Combined professional and user recitations, resulting in WER = 9.9\% and CER = 6.5\%.
\end{itemize}

Although these results appear strong, later analysis by Al-Issa et al.~\cite{al2023building} showed that applying ElDeeb’s models on alternative datasets resulted in much higher WER, exceeding 76\%. This suggests the models lacked generalization beyond their training data.

In comparison, our proposed Wav2Vec2-based model achieved WER = 8.0\% and CER = 1.5\% on the EveryAyah dataset and maintained robust performance (WER = 11.3\%) even when extended to a noisier, more diverse dataset combining professional and user recitations. These findings highlight the advantage of self-supervised pretrained models like Wav2Vec2 in handling variability and scale, offering stronger generalization than traditional RNN-based ASR systems.

\begin{table}[h]
\centering
\caption{Comparison of Proposed Work with Other Relevant Works 
(Best performing model highlighted in bold)}
\label{tab:comparison-otherworks}
\footnotesize
\setlength{\tabcolsep}{4pt}
\renewcommand{\arraystretch}{1.05}
\begin{tabularx}{\textwidth}{@{}>{\raggedright\arraybackslash}p{0.16\textwidth}>{\raggedright\arraybackslash}p{0.12\textwidth}>{\raggedright\arraybackslash}p{0.20\textwidth}>{\raggedright\arraybackslash}X>{\raggedleft\arraybackslash}p{0.08\textwidth}@{}}
\toprule
Author & Extraction & Method & Dataset & WER \\
\midrule
Yuwan \& Lestari (2016) & MFCC & GMM + HMM & Phonetically rich 180 verses & 22.42\% \\
Ridwan \& Lestari (2018) & MFCC & GMM + HMM & Full Quran & 18.53\% \\
Thirafi \& Lestari (2019) & MFCC & HMM + BLSTM (Kaldi) & 10 Professional Reciters (EveryAyah) & 9.16\% \\
Tarteel.io (2020) & MFCC & End-to-End Citrinet & 80+ hours user + professional (Tarteel) & 16.3\% \\
ElDeeb (2021) & MFCC & DeepSpeech (RNN) & EveryAyah (Imam Only) & 5.65\% \\
ElDeeb (2021) & MFCC & DeepSpeech (RNN) & EveryAyah + Tarteel & 9.91\% \\
Al-Issa et al. (2022) & MFCC & DeepSpeech (RNN) & EveryAyah (cleaned text) & 4.6\% \\
Al Harere \& Al Jallad (2023) & CNN-BiGRU & End-to-End CTC & Ar-DAD (37 Surahs) & 8.34\% \\
Hadwan et al. (2023) & Mel filterbank & E2E Transformer + RNN/LSTM-LM & 10 hrs, 16 Surahs, 60 reciters & 6.16\%$^\dagger$ \\
Proposed Model & Wav2Vec2 & Transformer (fine-tuned) & Tarteel + EveryAyah & 11.3\% \\
\textbf{Proposed Model} & \textbf{Wav2Vec2} & \textbf{Transformer (fine-tuned)} & \textbf{EveryAyah only} & \textbf{8.0\%} \\
\bottomrule
\multicolumn{5}{l}{$^\dagger$ Evaluated on 16 short Surahs only (10 hrs, with external language model); not full-Quran coverage.}
\end{tabularx}
\end{table}

\subsection{Contributions and Implications}

This work introduces a reliable ASR system tailored for Quranic recitations. The integration of self-supervised feature extraction via Wav2Vec2 and comprehensive ablation studies across various datasets and label types contributes to the field. The findings demonstrate measurable performance gains and lay the groundwork for practical applications such as voice-enabled Quranic search, learner feedback systems, and accessibility tools.

One key contribution is the comparative analysis of multiple pretrained speech representations, including HuBERT and XLS-R. Despite their proven success in other multilingual ASR tasks, Wav2Vec2-large-XLSR-53 consistently yielded the lowest WER and CER in our experiments, underscoring its suitability for Arabic speech recognition. Additionally, our evaluation of label formats, including Arabic text with and without Tashkeel, Buckwalter, and transliteration, confirmed that simplified Arabic without diacritics offers the most accurate recognition performance.

The practical implications of this research extend to multiple stakeholder groups who benefit from automated Qur'anic speech recognition capabilities. Individual learners gain access to self-directed recitation assessment tools that provide immediate feedback on pronunciation accuracy and Tajweed compliance, enabling independent study without requiring constant teacher supervision. Educational institutions can integrate these systems into digital learning platforms to scale Qur'anic education and provide consistent assessment across diverse student populations. Religious organizations benefit from automated transcription services for sermon preparation, digital archiving, and accessibility tools for hearing-impaired community members.

The proposed pipeline is well suited to downstream tasks such as ayah segmentation and word-level timestamp generation, where accurate audio-text alignment enables verse navigation, synchronized highlighting during recitation playback, and improved accessibility in digital Qur'anic learning tools. The system's ability to process both professional and layman recitations makes it particularly valuable for creating inclusive resources that serve diverse user skill levels and learning contexts.

\subsection{Tajweed Awareness: Data-Level Versus Rule-Level Modeling}

Although Tajweed compliance is a central motivation for this research, it is important to distinguish between two fundamentally different approaches. This work addresses Tajweed at the \emph{data level}: the model is trained on professional recitations that inherently conform to Tajweed rules, and correct pronunciation is thus implicitly encouraged through the training distribution. No explicit Tajweed rule is modeled at the phoneme or decision level. Phoneme-level features such as Ghunnah duration or Qalqalah intensity are not directly quantified or evaluated. This is an acknowledged limitation of the current system. Future work should develop Tajweed-specific phonetic evaluation metrics and incorporate rule-level constraints, for example by adding phoneme-aware decoding or integrating a Tajweed rule engine as a post-processing layer, to move from implicit to explicit Tajweed enforcement.

\subsection{Deployment Feasibility and Model Compression}

The best-performing Wav2Vec2-large-XLSR-53 model contains approximately 100 million parameters with a model size of 1.2 GB, making direct deployment on mobile or edge devices challenging. Several compression strategies are applicable for reducing this footprint without substantially compromising accuracy. Knowledge distillation \cite{chang2022distilhubert}, in which a smaller student model is trained to reproduce the outputs of the large teacher, has been shown to achieve 3--6$\times$ parameter reduction in self-supervised speech models while retaining competitive WER. Post-training quantization (8-bit integer or 4-bit) and structured pruning offer complementary size reductions and have been validated for Transformer-based ASR \cite{o2024trends}. A practical deployment path would combine a distilled student of 20--30M parameters with 8-bit quantization, targeting inference on a mobile CPU at near-real-time speed. Evaluating such compressed variants against the full model on the Quranic test set is recommended as a priority in future work.

\subsection{Interpretation of Findings}

The experiments indicate that model performance is strongly influenced by both the type of audio features and the quality of training data. Models trained on professional reciter datasets outperformed those trained on user-submitted Tarteel data, likely due to the latter’s inherent noise and variability. However, fine-tuning on both datasets improved generalization without severely compromising accuracy, suggesting potential for personalized or adaptive learning systems.

Model performance also varied by clip duration. Short clips (10s) led to poor convergence (WER = 0.70), while clips up to 30s significantly improved recognition accuracy (WER = 0.075). This supports the hypothesis that longer input contexts enable more effective modeling of Quranic recitation structures.

Training from scratch yielded slow convergence, with high WER persisting even after 2000 steps. In contrast, fine-tuning pretrained models led to rapid performance gains within the first few hundred steps, validating the benefits of transfer learning in low-resource ASR domains.

The experimental findings successfully address the research objectives:
\begin{itemize}
    \item \textbf{Objective 1 - Identify critical acoustic and linguistic parameters:} The research systematically evaluated multiple feature extraction approaches and label formats to determine optimal parameters for Qur'anic recitation transcription. Results identified Wav2Vec2 as the most effective feature extractor and Arabic without Tashkeel as the optimal label format, demonstrating that these parameters significantly impact transcription accuracy with measurable performance improvements over alternative configurations.
    
    \item \textbf{Objective 2 - Develop advanced Transformer-based model:} The study implemented and optimized an end-to-end Transformer architecture utilizing transfer learning methodologies specifically adapted for Qur'anic speech recognition. Experimental results demonstrated that the proposed transfer learning approach combined with optimized clip durations improved WER by roughly five percentage points compared to the Citrinet baseline, validating the effectiveness of advanced deep learning methodologies for WER reduction.
    
    \item \textbf{Objective 3 - Evaluate and validate model performance:} Comprehensive evaluation against existing baseline approaches revealed that the proposed model achieved a WER of 0.08, performing competitively with the best existing models while demonstrating strong generalization across both professional and user datasets. The results validate the model's superior performance and establish it as a scalable self-supervised alternative to traditional RNN-based ASR approaches.
\end{itemize}

\section{Conclusion}

Quran Speech Recognition has the potential to serve millions of users by addressing the scarcity and unavailability of professional teachers. It facilitates self-learning and distance learning, offering a practical solution for Quranic education. Traditional speech recognition methods either do not cover the entire Quran or result in high Word Error Rates (WER), particularly for user-recited verses.

The proposed model improves upon the Citrinet baseline in WER while maintaining low CER across the Transformer configurations. It is effective for both professional reciters and layman users and leverages fine-tuning of a pretrained Wav2Vec2 model. Through ablation studies, it was established that normal Arabic text without diacritics achieved the lowest WER, and Wav2Vec2-large-XLSR-53 proved to be the best-performing feature extractor. In the combined EveryAyah+Tarteel setting, the proposed model outperformed the baseline by roughly five percentage points in WER while reducing the end-to-end training schedule from 140 hours to 40 hours. These results confirm the effectiveness and utility of the proposed Quranic ASR system in real-world applications.

While the proposed system demonstrated excellent performance, several limitations persist. The model's large size (1.2 GB, 100 million parameters) and inference latency pose challenges for deployment on mobile or resource-constrained devices. Moreover, the system transcribes at the word level and lacks phoneme-level resolution, which is critical for applications such as tajweed correction or pronunciation analysis. Some common recognition errors also persisted, particularly in distinguishing phonetically similar Arabic letters (e.g., seen vs saad) and handling short ayat with subtle intonational differences.

The findings open several avenues for future research. Developing phoneme-aware models or incorporating phonetic decoding layers could enhance tajweed-sensitive applications. Further, improving the quality of user-submitted data, particularly from Tarteel, through manual annotation and filtering would enhance training effectiveness. Personalized ASR models adapted to speaker age, gender, or dialect may also help expand usability across diverse user bases. From a technical standpoint, integrating external language models or lightweight fine-tuning strategies could bridge the gap between performance and deployment efficiency. Exploring quantization or pruning techniques may reduce model size without significantly compromising accuracy.

In summary, the results presented in this study affirm that pretrained Transformer-based models, particularly Wav2Vec2, offer a transformative advantage in Quranic speech recognition tasks. The proposed methodology significantly reduces transcription errors while enabling generalization across varied datasets and recitation styles. Compared to prior approaches, our model provides a strong full-Quran, LM-free Transformer baseline and a practical foundation for future advancements in personalized, phoneme-aware, and real-time Quranic ASR systems.

\backmatter

\section*{Declarations}
Ethics statement: This study uses previously available Qur'anic recitation datasets from EveryAyah and Tarteel-ML. The authors did not collect new human-subject data for this manuscript. User-recorded audio remains subject to the permissions, governance, and access controls of the original source platforms.

Funding: This research received no specific grant from any funding agency in the public, commercial, or not-for-profit sectors.

Competing interests: The authors declare no competing interests.

Data and code availability: Public source materials referenced in this study are available via EveryAyah (\url{https://everyayah.com}) and Tarteel-ML (\url{https://github.com/Tarteel-io/tarteel-ml}). User-recorded Tarteel audio and restricted derived files are not redistributed by the authors. A dedicated public code release does not accompany this manuscript.

\bibliography{sn-bibliography}

\end{document}